\documentclass[10pt,twocolumn,letterpaper]{article}

\usepackage[pagenumbers]{cvpr}

\usepackage{graphicx}
\usepackage{amsmath}
\usepackage{amssymb}
\usepackage{booktabs}
\usepackage{url}

\usepackage{utfsym}
\usepackage{pifont}
\usepackage{soul}
\usepackage{dsfont}

\usepackage{subcaption}
\usepackage{caption}

\usepackage[pagebackref,breaklinks,colorlinks]{hyperref}

\usepackage[capitalize]{cleveref}
\crefname{section}{Sec.}{Secs.}
\Crefname{section}{Section}{Sections}
\Crefname{table}{Table}{Tables}
\crefname{table}{Tab.}{Tabs.}

\makeatletter
\newcommand{\thickhline}{%
  \noalign {\ifnum 0=`}\fi \hrule height 1pt
  \futurelet \reserved@a \@xhline
}
\makeatother

\def\cvprPaperID{*****}
\def\confName{CVPR}
\def\confYear{2025}

\begin{document}

\title{A VideoMAE-v2 Approach to Zero-Shot Traffic Accident Anticipation}

\author{
Team BUPT MIC Lab \\
Siyuan Li$^{1}$,
Xiaoyang Bi$^{1}$, and
Mengshi Qi$^{1}$\thanks{Corresponding author}\\
$^{1}$State Key Laboratory of Networking and Switching Technology \\
$^{2}$Beijing University of Posts and Telecommunications \\
{\tt\small \{lisiyuan0818, bxy, qms\}@bupt.edu.cn}
}

\maketitle

\begin{abstract}
Traffic accident anticipation---predicting the likelihood of an
imminent collision at every frame of a dashcam video---is
safety-critical yet difficult to scale, because collecting in-domain
annotated accident footage for every deployment scenario is
prohibitively expensive.  We study this task under a \emph{zero-shot}
setting where no target-domain training data is available: the model
must learn exclusively from a publicly available binary-labelled
driving-accident dataset and generalise to unseen dashcam footage.
We propose a framework that bridges the gap between the
frame-level temporal risk estimation task and coarsely labelled
binary accident datasets by coupling a VideoMAE-v2 backbone with a
per-frame prediction head under a sliding-window protocol.
Our method achieves \textbf{2nd place} in the 2026 CVPR@AUTOPILOT
Zero-Shot Accident Anticipation competition~\cite{AUTOPILOT-COG}. Code is available at \url{https://github.com/TimeSouth/zero-shot-taa-solution}.

\end{abstract}

\section{Introduction}
\label{sec:intro}

Traffic accident anticipation (TAA) aims to estimate, at every frame
of a dashcam video, the probability that a collision or near-miss will
occur in the immediate future. Accurate and timely frame-level risk
prediction is a prerequisite for driver-assistance alerts and
autonomous emergency braking~\cite{bao2020uncertainty,karim2022dsta}, alongside other safety-critical perception tasks in robust autonomous driving and 3D scene understanding~\cite{liao2026improving,ye2025safedriverag,lv2025t2sg,zhu2023unsupervised,wang2022rgb,wang2024rdfc}.
Prior work has achieved encouraging results when in-domain annotated
data is available~\cite{bao2020uncertainty,karim2022dsta,fang2022dada,yao2022dota};
however, collecting and labelling dashcam accident footage for every
target domain is prohibitively expensive.  A more practical paradigm
is \emph{zero-shot} TAA, in which the model is trained solely on
publicly available driving-accident data and must generalise to
previously unseen dashcam distributions without any target-domain
supervision.

Under this zero-shot setting, two principal technical challenges
arise.  First, a \emph{task--data granularity mismatch}: TAA requires
dense, frame-level risk scores over clips of a fixed temporal format
(\eg, 150 frames at 30\,fps), yet publicly available accident datasets
provide only coarse binary labels (accident vs.\ normal).  Bridging
this granularity gap during training is essential for meaningful
per-frame supervision.
Second, a \emph{long-clip / short-window modelling} problem: while advanced video representation and multimodal learning have driven significant progress across various visual tasks (e.g., action assessment, temporal localization, video retrieval, and commonsense reasoning)~\cite{dc-sam,disentangled,qi2025explainable,qi2026towards,qi2025action,yun2024weakly,qi2021semantics,qi2020few,qi2019sports}, state-of-the-art
video encoders such as VideoMAE-v2~\cite{wang2023videomaev2} accept
only 16 frames per forward pass, substantially fewer than the full
clip length, necessitating a principled window-to-clip aggregation
mechanism.

We address the first challenge by restructuring a publicly available binary-labelled
accident dataset into temporally standardised clips with explicit
positive--negative balancing, so that per-frame supervision can be
derived from clip-level annotations (\cref{subsec:data}).  We address
the second by coupling a VideoMAE-v2 backbone with a per-frame prediction
head under a sliding-window protocol, and recovering dense clip-level
risk sequences via overlap averaging and temporal interpolation
(\cref{subsec:train}).  Furthermore, we observe that the source-trained model
suffers from a systematic \emph{posterior shift} when applied to the
target domain: the relative ranking of frame-level predictions is
well-preserved, yet the predicted confidence distribution is
systematically compressed, preventing effective decision-making.  We
address this with a lightweight, training-free test-time domain
adaptation module that leverages task-specific temporal priors and
the model's own prediction-distribution statistics to recalibrate
the output without any target-domain labels (\cref{subsec:post}).

Our method achieves \textbf{2nd place} in the 2026 CVPR@AUTOPILOT
Zero-Shot Accident Anticipation competition~\cite{AUTOPILOT-COG}.  Our design choices are
further validated by two ablations (\cref{sec:exp}) demonstrating that two seemingly attractive
alternatives---\emph{late-window classification} and
\emph{cross-dataset training without class balancing}---both lead to
significant performance degradation.

\section{Method}
\label{sec:method}

\subsection{Task Definition and Notation}
\label{subsec:overview}

Given a dashcam video clip of $T$ frames, traffic accident anticipation
(TAA) requires producing a per-frame risk sequence
$\{p_t\}_{t=1}^{T}$, where $p_t\!\in\![0,1]$ denotes the estimated
probability that a collision or near-miss will occur \emph{after}
frame~$t$ within the same clip.  In the zero-shot setting studied
here, no target-domain training data is available; the model is
trained on an external source dataset (\eg, Nexar~\cite{nexar2025zerotaa}) in which each video carries only
a clip-level binary label $y\!\in\!\{0,1\}$ (accident vs.\ normal)
and the event onset time~$\tau$.  During training, we derive
frame-level supervision $y_t\!\in\!\{0,1\}$ from these clip-level
annotations (detailed in \cref{subsec:data}).

Our pipeline consists of three stages: (i)~\emph{dataset curation}
(\cref{subsec:data}), which restructures the source data into
temporally standardised clips with balanced class distribution;
(ii)~\emph{model training and inference}
(\cref{subsec:train}), which fine-tunes a
video encoder to predict $\{p_t\}$ via a sliding-window protocol; and
(iii)~\emph{test-time domain adaptation} (\cref{subsec:post}), which
recalibrates the output distribution at test time by exploiting
task-specific temporal priors and the model's own prediction-distribution
statistics, without requiring any target-domain labels.

\subsection{Dataset Construction}
\label{subsec:data}

\noindent\textbf{Source.}~We use the public Nexar Crash / Dashcam
Collision Prediction dataset~\cite{nexar2025zerotaa} as the sole training and validation
source.  Each video is annotated with a clip-level label (collision /
near-miss vs.\ normal driving) and the time-of-event $\tau$.

\noindent\textbf{Clip Splitting.}~Each competition~\cite{AUTOPILOT-COG} test clip is exactly
150 frames at 30\,fps, so we slice every Nexar video to match.  For
positive videos we discard the last 2\,s before $\tau$, then slide a 5\,s window over the remaining footage, so the model never observes
the collision moment itself.  For negative videos we slide directly
and set $\tau\!=\!250$ (well outside the clip) so that the
time-weighted CE term reduces to plain CE.  Frame-level labels follow
$y_t\!=\!1$ in positive clips and $y_t\!=\!0$ otherwise.

\noindent\textbf{Class Balancing.}~A naïve sliding-window split is
heavily biased toward negative clips, on which a CE objective collapses
to the trivial ``always-negative'' minimiser.  We therefore apply a
tiered balancing scheme that (i) keeps every positive clip,
(ii) keeps every negative clip cut from a positive video---these
provide informative \emph{pre-/post-event} context, and (iii)
sub-samples clips from the purely-negative videos breadth-first
(at most three per video) until the positive : negative ratio is
roughly 1:1.  Capping per-video rather than uniformly sub-sampling
preserves diversity in scenes, weather and camera setups, which we
found to be important for cross-domain generalisation.

\noindent\textbf{Input Preprocessing.}~We down-sample 30\,fps to
10\,fps (frame interval of 3), so a 5\,s clip becomes 50
frames; we then extract overlapping 16-frame windows with stride 4 to
match the VideoMAE input.  Frames are resized to $224\!\times\!224$ and
ImageNet-normalised.  Training augments with horizontal flip
($p\!=\!0.5$) and brightness/contrast jitter ($p\!=\!0.3$).

\subsection{Training and Inference}
\label{subsec:train}

\noindent\textbf{Architecture.}~We adopt VideoMAE-v2
Base~\cite{wang2023videomaev2,tong2022videomae,he2022mae}
as the video encoder.  Each 16-frame window
$X\!\in\!\mathbb{R}^{16\times 3\times 224\times 224}$ is encoded into
a set of spatio-temporal tokens
$Z\!\in\!\mathbb{R}^{N\times D}$.  On top of the encoder we attach
a lightweight per-frame classification head that
fuses a global mean-pooled feature with temporally upsampled token
features, producing per-frame risk probabilities
$\{p_t^{(w)}\}_{t=1}^{16}\!\in\![0,1]$ for each window.  The entire
model is trained end-to-end.

\noindent\textbf{Training objective.}~For window samples with frame
label $y_t\!\in\!\{0,1\}$, time-to-event $\tau$ and global frame
index $t$, we minimise
\begin{equation}
\mathcal{L} = \tfrac{1}{2}\Big(
  \mathbb{E}_{y_t=1}\!\big[\,e^{-\max(0,\,(\tau-t-1)/\text{fps})}\!\cdot\!\mathrm{CE}\big]
  + \mathbb{E}_{y_t=0}[\mathrm{CE}]
\Big)
\end{equation}
This is the discrete ``Exp-Loss'' commonly used for accident
anticipation~\cite{bao2020uncertainty}: it down-weights frame
predictions that are far from the event so that the model is
encouraged to be confident only when the accident is temporally
imminent, while still being penalised by full CE elsewhere.

\noindent\textbf{Window-to-clip aggregation.}~For each 150-frame test
clip we (i)~slide a 16-frame window with stride 1 across the
10\,fps-sampled sequence, (ii)~run the network on every window to
obtain $\{p_t^{(w)}\}$, (iii)~for every sampled frame position average
the predictions of all windows that cover it (\emph{overlap
averaging}), and (iv)~linearly interpolate the resulting sparse
probability vector back to all 150 frame indices (\emph{temporal
interpolation}).  Step~(iv) is necessary because the model only
outputs predictions at 10\,fps-sampled positions, while the evaluation
requires 30\,fps frame-level scores.

\subsection{Test-Time Domain Adaptation}
\label{subsec:post}

We observe that the source-trained model suffers from a
\emph{posterior shift} on the target domain: frame-level risk rankings
are well-preserved, yet the absolute confidence scale is systematically
compressed, a common artefact in unsupervised domain
adaptation.  We address this with a training-free, three-component
test-time adaptation module.

\noindent\textbf{Temporal confidence aggregation.}~In the TAA task,
frames closer to the end of a clip carry stronger causal evidence
about an impending event, since the accident---if any---occurs
shortly after the clip ends.  Motivated by this \emph{temporal
saliency prior}, we aggregate the per-frame predictions into a single
clip-level confidence anchor $p_{\text{final}}$ by retaining the
last-frame prediction, which empirically exhibits the highest
signal-to-noise ratio among all frame positions.

\noindent\textbf{Temporal risk prior reconstruction.}~Accident risk
possesses a well-known \emph{monotonicity prior}: the probability of
an imminent collision increases monotonically as the temporal distance
to the event decreases.  We exploit this inductive bias to
reconstruct a dense, physically plausible frame-level risk curve from
the clip-level anchor.  Specifically, we parameterise the
reconstruction as
\begin{equation}
p_t = p_{\text{start}} + (\beta\,p_{\text{final}}\!-\!p_{\text{start}})
\!\cdot\!h(t/T;\,k,\mu),
\end{equation}
where $h$ is a monotonically increasing temporal basis function that
blends exponential and logarithmic schedules, $k$ and $\mu$ are shape
parameters estimated from the second-order statistics of the raw
curve, $\beta < 1$ is a head-room factor ensuring strict monotonicity
below the end anchor, and $p_{\text{start}}$ is the initial risk
level.  This step effectively \emph{propagates} the model's
high-confidence late-segment signal backward along the temporal axis
under a physically motivated constraint.

\noindent\textbf{Prediction distribution alignment.}~Even after
temporal reconstruction, the predicted confidence distribution on the
target domain remains systematically shifted relative to the source
domain's decision boundary.  Since no target-domain labels are
available, we perform \emph{unsupervised recalibration} using only
the moments of the model's own prediction distribution on the target
set.  Concretely, we compute the empirical median $m$ of
$\{p_{\text{final}}\}$ across all target clips as a domain-specific
statistic, and apply an order-preserving mapping that aligns the
predicted distribution to the expected decision boundary.  This
procedure is analogous to test-time batch normalisation in
unsupervised domain adaptation: it shifts the activation statistics
to match the target distribution without retraining.  Crucially,
because the only target-set quantity used is a moment of the model's
own output distribution, the zero-shot constraint is fully respected.

\section{Experiments}
\label{sec:exp}

\subsection{Implementation Details}
The full model (87.1\,M parameters) is fine-tuned with
AdamW~\cite{loshchilov2019adamw}, learning rate $1\!\times\!10^{-5}$
for the backbone and $1\!\times\!10^{-4}$ for the head, weight decay
$5\!\times\!10^{-4}$, batch size 3 with 8-step gradient accumulation
(effective batch 24), 20 epochs, StepLR ($\gamma\!=\!0.1$ every 5
epochs), gradient clipping with max norm $5.0$, and automatic mixed
precision (AMP).  Training takes approximately 17 hours on a single
NVIDIA V100 GPU.  We select the submitted checkpoint based on the
validation AP, AUC, and mTTA.

\noindent\textbf{Model scale.}~The VideoMAE-v2 Base backbone has
86.2\,M parameters; the per-frame head adds 0.9\,M, totalling
87.1\,M trainable parameters.

\noindent\textbf{Test-time adaptation hyper-parameters.}~On the source
validation set, the raw model produces median
$p_{\text{final}}\!\approx\!0.34$ and clip-mean $\approx\!0.05$.
The temporal risk prior uses $k\!\in\![7,15]$, $\mu\!\in\![0.2,0.8]$, and
head-room factor $\beta\!=\!0.998$.  The distribution alignment step
maps clips in $[m, 0.5)$ order-preservingly into $[0.505, 0.7]$.

\subsection{Evaluation Protocol}
We follow the official evaluation protocol of the CVPR@AUTOPILOT Zero-Shot Accident Anticipation competition~\cite{AUTOPILOT-COG}. The evaluation measures detection quality via frame-level Average Precision (AP) and Area Under the ROC Curve (AUC), while early and stable risk awareness is assessed via Time-To-Accident (TTA) and Stable TTA (STTA). Higher AP and AUC indicate better detection capability, whereas larger TTA metrics reflect earlier anticipation. On our held-out Nexar~\cite{nexar2025zerotaa} validation split, we report the AP and AUC metrics. Additionally, we present the competition's official weighted composite score evaluated on the unseen test set, denoted as ``Private LB'' in our results.

\subsection{Results}

Beyond the final pipeline of \cref{sec:method}, we ran two experiments to test design choices that, \emph{a priori}, looked
promising but turned out to hurt the score.  Both are reported here
because their negative results directly motivate decisions in our
final method.  In both, the \emph{only} change from the final pipeline
is the one explicitly stated; everything else is held constant.

\noindent\textbf{Late-Window Classification.}~Instead of feeding the
model a 16-frame window over the full 5\,s clip, we extract only the
\emph{last 2\,s} of every clip, uniformly sub-sample 16 frames from
this trailing window, and produce a single clip-level binary decision
(rather than a per-frame risk sequence).  The clip-level probability
is then replicated across all 150 output positions.  Trained with the
same backbone and class-balanced split as the proposed model, this
variant reaches a validation $\text{AP}=0.66$ and $\text{AUC}=0.69$
(\cref{tab:exp}, row B), clearly below the proposed per-frame model.
On the private leaderboard the gap widens to $-0.027$ with respect to
our final submission.  The underlying reason is that, by replicating a
single scalar across all 150 positions, the late-window output has no
temporal resolution by construction, so the post-processor of
\cref{subsec:post}---which is order-preserving---cannot rebuild a
rising risk curve from a flat one.  Empirically, this confirms that
\emph{discarding the first 3\,s of context costs more than the simpler
late-time supervision saves}; we therefore retain the full 5\,s input
plus a per-frame head.

\noindent\textbf{Cross-Dataset Mixed Training without Class
Balancing.}~Motivated by the scarcity of accident clips in Nexar, we
augment training with a second public accident
dataset~\cite{yao2022dota} and use the union as the new training set.
Crucially, in this experiment we \emph{bypass the class-balancing step
of} \cref{subsec:data}: every positive clip from both sources is kept,
and only Nexar's purely-negative videos are sub-sampled.  The resulting
training set contains substantially more positives than negatives.
The model trained under this regime degenerates into an
``always-positive'' predictor: on validation it still attains
$\text{AP}=0.78$ (\cref{tab:exp}, row C), because AP on an imbalanced
set rewards any model that tends to predict the majority class, but
the corresponding $\text{AUC}=0.60$---close to the $0.5$ chance
level---reveals that the model has essentially lost the ability to
rank positives above negatives.  The private-leaderboard score drops
to $2.35$, a full $-0.108$ gap below our final submission.  This is a
textbook consequence of class imbalance at the dataset level: when the
marginal $P(y\!=\!1)$ in training is pushed close to one, the
cross-entropy minimiser is the constant-positive predictor, and the
time-weighting term in the Exp-Loss---which only re-scales the
\emph{positive} loss---provides no counterweight.  It is for exactly
this reason that we make dataset-construction-time class balancing a
first-class component of our pipeline rather than a tunable
hyper-parameter.

\begin{table}[t]
\centering
\small
\setlength{\tabcolsep}{4pt}
\begin{tabular}{clccc}
\toprule
& Method & AP & AUC & Private LB \\
\midrule
A & Proposed                & $\textbf{0.83}$ & $\textbf{0.86}$ & $\mathbf{2.46234}$ \\
B & Late-Window             & $0.66$ & $0.69$ & $2.43530$ \\
C & Mixed w/o balance       & $0.78$ & $0.60$ & $2.35424$ \\
\bottomrule
\end{tabular}
\caption{Comparisons on the Nexar validation split.  ``AP''
and ``AUC'' are computed on the held-out Nexar validation set, with
AP reported as the standard average precision in $[0,1]$.
``Private LB'' reports the corresponding submission's score on the
official Kaggle \emph{private} leaderboard of the
CVPR@AUTOPILOT Zero-Shot Accident Anticipation competition, where
higher is better.}
\label{tab:exp}
\end{table}

\section{Conclusion}
We presented a zero-shot traffic accident anticipation framework that
couples a VideoMAE-v2 sliding-window classifier with class-balanced
dataset construction and a training-free test-time domain adaptation
module.  Ablation studies confirm that both full-context per-frame
modelling and dataset-level class balancing are essential to avoid
performance collapse.  The system achieves 2nd place in the 2026
CVPR@AUTOPILOT Zero-Shot Accident Anticipation competition~\cite{AUTOPILOT-COG}.

{\small
\bibliographystyle{ieee_fullname}
\bibliography{egbib}

@article{tong2022videomae,
  title={Videomae: Masked autoencoders are data-efficient learners for self-supervised video pre-training},
  author={Tong, Zhan and Song, Yibing and Wang, Jue and Wang, Limin},
  journal={Advances in neural information processing systems},
  volume={35},
  pages={10078--10093},
  year={2022}
}

@inproceedings{wang2023videomaev2,
  title={Videomae v2: Scaling video masked autoencoders with dual masking},
  author={Wang, Limin and Huang, Bingkun and Zhao, Zhiyu and Tong, Zhan and He, Yinan and Wang, Yi and Wang, Yali and Qiao, Yu},
  booktitle={Proceedings of the IEEE/CVF conference on computer vision and pattern recognition},
  pages={14549--14560},
  year={2023}
}

@inproceedings{he2022mae,
  title={Masked autoencoders are scalable vision learners},
  author={He, Kaiming and Chen, Xinlei and Xie, Saining and Li, Yanghao and Doll{\'a}r, Piotr and Girshick, Ross},
  booktitle={Proceedings of the IEEE/CVF conference on computer vision and pattern recognition},
  pages={16000--16009},
  year={2022}
}

@article{fang2022dada,
  title={DADA: Driver attention prediction in driving accident scenarios},
  author={Fang, Jianwu and Yan, Dingxin and Qiao, Jiahuan and Xue, Jianru and Yu, Hongkai},
  journal={IEEE transactions on intelligent transportation systems},
  volume={23},
  number={6},
  pages={4959--4971},
  year={2021},
  publisher={IEEE}
}

@article{yao2022dota,
  title={DoTA: Unsupervised detection of traffic anomaly in driving videos},
  author={Yao, Yu and Wang, Xizi and Xu, Mingze and Pu, Zelin and Wang, Yuchen and Atkins, Ella and Crandall, David J},
  journal={IEEE transactions on pattern analysis and machine intelligence},
  volume={45},
  number={1},
  pages={444--459},
  year={2022},
  publisher={IEEE}
}

@inproceedings{bao2020uncertainty,
  title={Uncertainty-based traffic accident anticipation with spatio-temporal relational learning},
  author={Bao, Wentao and Yu, Qi and Kong, Yu},
  booktitle={Proceedings of the 28th ACM International Conference on Multimedia},
  pages={2682--2690},
  year={2020}
}

@article{karim2022dsta,
  title={A dynamic spatial-temporal attention network for early anticipation of traffic accidents},
  author={Karim, Muhammad Monjurul and Li, Yu and Qin, Ruwen and Yin, Zhaozheng},
  journal={IEEE Transactions on Intelligent Transportation Systems},
  volume={23},
  number={7},
  pages={9590--9600},
  year={2022},
  publisher={IEEE}
}

@article{loshchilov2019adamw,
  title={Decoupled weight decay regularization},
  author={Loshchilov, Ilya and Hutter, Frank},
  journal={arXiv preprint arXiv:1711.05101},
  year={2017}
}

@misc{nexar2025zerotaa,
  author       = {Daniel C. Moura and Shizhan Zhu and Orly Zvitia},
  title        = {Nexar Dashcam Crash Prediction Challenge},
  year         = {2025},
  howpublished = {\url{https://kaggle.com/competitions/nexar-collision-prediction}},
  note         = {Kaggle}
}

@misc{AUTOPILOT-COG,
  author       = {{AUTOPILOT-COG}},
  title        = {Zero-shot Accident Anticipation},
  year         = {2026},
  howpublished = {\url{https://kaggle.com/competitions/zero-shot-taa}},
  note         = {Kaggle}
}

@ARTICLE{dc-sam,
  author={Qi, Mengshi and Zhu, Pengfei and Li, Xiangtai and Bi, Xiaoyang and Qi, Lu and Ma, Huadong and Yang, Ming-Hsuan},
  journal={IEEE Transactions on Pattern Analysis and Machine Intelligence}, 
  title={DC-SAM: In-Context Segment Anything in Images and Videos via Dual Consistency}, 
  year={2026},
  volume={48},
  number={4},
  pages={4642-4656},
  doi={10.1109/TPAMI.2025.3646919}}

@ARTICLE{ disentangled,
  author={Qi, Mengshi and Lv, Changsheng and Ma, Huadong},
  journal={IEEE Transactions on Pattern Analysis and Machine Intelligence}, 
  title={Robust Disentangled Counterfactual Learning for Physical Audiovisual Commonsense Reasoning}, 
  year={2026},
  volume={48},
  number={3},
  pages={2514-2527},
  doi={10.1109/TPAMI.2025.3627224}}

@article{qi2025explainable,
  title={Explainable Action Form Assessment by Exploiting Multimodal Chain-of-Thoughts Reasoning},
  author={Qi, Mengshi and Wu, Yeteng and Zhang, Xianlin and Ma, Huadong},
  journal={arXiv preprint arXiv:2512.15153},
  year={2025}
}

@inproceedings{qi2026towards,
  title={Towards Balanced Multi-Modal Learning in 3D Human Pose Estimation},
  author={Qi, Mengshi and Peng, Jiaxuan and Zhang, Xianlin and Ma, Huadong},
  booktitle={Proceedings of the IEEE/CVF Conference on Computer Vision and Pattern Recognition},
  pages={21231--21241},
  year={2026}
}

@article{qi2025action,
  title={Action quality assessment via hierarchical pose-guided multi-stage contrastive regression},
  author={Qi, Mengshi and Ye, Hao and Peng, Jiaxuan and Ma, Huadong},
  journal={IEEE Transactions on Image Processing},
  year={2025},
  publisher={IEEE}
}

@inproceedings{yun2024weakly,
  title={Weakly-supervised temporal action localization by inferring salient snippet-feature},
  author={Yun, Wulian and Qi, Mengshi and Wang, Chuanming and Ma, Huadong},
  booktitle={Proceedings of the AAAI conference on artificial intelligence},
  volume={38},
  number={7},
  pages={6908--6916},
  year={2024}
}

@article{qi2021semantics,
  title={Semantics-aware spatial-temporal binaries for cross-modal video retrieval},
  author={Qi, Mengshi and Qin, Jie and Yang, Yi and Wang, Yunhong and Luo, Jiebo},
  journal={IEEE Transactions on Image Processing},
  volume={30},
  pages={2989--3004},
  year={2021},
  publisher={IEEE}
}

@inproceedings{qi2020few,
  title={Few-shot ensemble learning for video classification with slowfast memory networks},
  author={Qi, Mengshi and Qin, Jie and Zhen, Xiantong and Huang, Di and Yang, Yi and Luo, Jiebo},
  booktitle={Proceedings of the 28th ACM international conference on multimedia},
  pages={3007--3015},
  year={2020}
}

@article{qi2019sports,
  title={Sports video captioning via attentive motion representation and group relationship modeling},
  author={Qi, Mengshi and Wang, Yunhong and Li, Annan and Luo, Jiebo},
  journal={IEEE Transactions on Circuits and Systems for Video Technology},
  volume={30},
  number={8},
  pages={2617--2633},
  year={2019},
  publisher={IEEE}
}

@inproceedings{liao2026improving,
  title={Improving Batch Normalization with Test-Time Adaptation for Robust Object Detection in Self-Driving},
  author={Liao, Dacheng and Qi, Mengshi and Liu, Liang and Ma, Huadong},
  booktitle={Proceedings of the AAAI Conference on Artificial Intelligence},
  volume={40},
  number={9},
  pages={6925--6933},
  year={2026}
}

@inproceedings{ye2025safedriverag,
  title={Safedriverag: Towards safe autonomous driving with knowledge graph-based retrieval-augmented generation},
  author={Ye, Hao and Qi, Mengshi and Liu, Zhaohong and Liu, Liang and Ma, Huadong},
  booktitle={Proceedings of the 33rd ACM International Conference on Multimedia},
  pages={11170--11178},
  year={2025}
}

@inproceedings{zhu2023unsupervised,
  title={Unsupervised self-driving attention prediction via uncertainty mining and knowledge embedding},
  author={Zhu, Pengfei and Qi, Mengshi and Li, Xia and Li, Weijian and Ma, Huadong},
  booktitle={Proceedings of the IEEE/CVF international conference on computer vision},
  pages={8558--8568},
  year={2023}
}

@inproceedings{lv2025t2sg,
  title={T2sg: Traffic topology scene graph for topology reasoning in autonomous driving},
  author={Lv, Changsheng and Qi, Mengshi and Liu, Liang and Ma, Huadong},
  booktitle={Proceedings of the Computer Vision and Pattern Recognition Conference},
  pages={17197--17206},
  year={2025}
}

@inproceedings{wang2022rgb,
  title={Rgb-depth fusion gan for indoor depth completion},
  author={Wang, Haowen and Wang, Mingyuan and Che, Zhengping and Xu, Zhiyuan and Qiao, Xiuquan and Qi, Mengshi and Feng, Feifei and Tang, Jian},
  booktitle={Proceedings of the ieee/cvf conference on computer vision and pattern recognition},
  pages={6209--6218},
  year={2022}
}

@article{wang2024rdfc,
  title={RDFC-GAN: RGB-depth fusion CycleGAN for indoor depth completion},
  author={Wang, Haowen and Che, Zhengping and Yang, Yufan and Wang, Mingyuan and Xu, Zhiyuan and Qiao, Xiuquan and Qi, Mengshi and Feng, Feifei and Tang, Jian},
  journal={IEEE Transactions on Pattern Analysis and Machine Intelligence},
  volume={46},
  number={11},
  pages={7088--7101},
  year={2024},
  publisher={IEEE}
}
}

\end{document}